\documentclass{article}

\usepackage[final,nonatbib]{neurips_2024}

\usepackage[utf8]{inputenc} 
\usepackage[T1]{fontenc}    
\usepackage{hyperref}       
\usepackage{url}            
\usepackage{booktabs}       
\usepackage{amsfonts}       
\usepackage{nicefrac}       
\usepackage{microtype}      
\usepackage{xcolor}         
\usepackage{bm}
\usepackage{xspace}
\usepackage{graphicx}
\usepackage{animate}
\usepackage{multirow}
\usepackage{xurl}
\usepackage{amsmath}
\usepackage{enumitem}
\usepackage{xcolor}
\usepackage{wrapfig}

\newcommand{\name}{VASA-1\xspace}

\graphicspath{{figures/}}

\title{\name: Lifelike Audio-Driven Talking Faces\\ Generated in Real Time}

\author{Microsoft Research Asia}

\author{
	\!\!\!\!\!\! Sicheng Xu\thanks{: Equal contributions. $^{\dag}$: Corresponding author. See the contribution statement section for contributions.
	}\ ~\\
	\!\!\!\!Microsoft Research Asia  \\
	\texttt{\!\!\!\!\!\!sichengxu@microsoft.com\!\!}
	\And \!\!\!\!Guojun Chen${^*}$ \\
	\!\!\!\!Microsoft Research Asia  \\
	\texttt{\!\!\!\!guoch@microsoft.com}
	\And Yu-Xiao Guo${^*}$ \\
	Microsoft Research Asia  \\
	\texttt{yuxgu@microsoft.com}
	\And \ \ \ Jiaolong Yang${^{*\ \!\dag}}$ \\
	Microsoft Research Asia  \\
	\texttt{jiaoyan@microsoft.com}
	\And Chong Li\ ~ \ \\
	Microsoft Research Asia  \\
	\texttt{chol@microsoft.com}  
	\And Zhenyu Zang\\
	Microsoft Research Asia  \\
	\texttt{\!\!\!\!zhenyuzang@microsoft.com\!\!\!\!} 
	\And Yizhong Zhang \\
	Microsoft Research Asia  \\
	\texttt{yizzhan@microsoft.com\!} 
	\And Xin Tong \\
	 Microsoft Research Asia  \\
	\texttt{xtong@microsoft.com}
	\And Baining Guo \\
	Microsoft Research Asia  \\
	\texttt{bainguo@microsoft.com}
}

\begin{document}

\maketitle

\begin{abstract}
We introduce VASA, a framework for generating lifelike talking faces with appealing visual affective skills (VAS) given a single static image and a speech audio clip. Our premiere model, \name, is capable of not only generating lip movements that are exquisitely synchronized with the audio, but also producing a large spectrum of facial nuances and natural head motions that contribute to the perception of authenticity and liveliness. 
The core innovations include a diffusion-based holistic facial dynamics and head movement generation model that works in a face latent space, and the development of such an expressive and disentangled face latent space using videos.
Through extensive experiments including evaluation on a set of new metrics, we show that our method significantly outperforms previous methods along various dimensions comprehensively. Our method delivers high video quality with realistic facial and head dynamics and also supports the online generation of 512$\times$512 videos at up to 40 FPS with negligible starting latency.
It paves the way for real-time engagements with lifelike avatars that emulate human conversational behaviors. 
Project webpage: \url{https://www.microsoft.com/en-us/research/project/vasa-1/}

\end{abstract}

\section{Introduction}

In the realm of multimedia and communication, the human face is not just a visage but a dynamic canvas, where every subtle movement and expression can articulate emotions, convey unspoken messages, and foster empathetic connections. 
The emergence of AI-generated talking faces offers a window into a future where technology amplifies the richness of human-human and human-AI interactions. Such technology holds the promise of enriching digital communication~\cite{wang2021one,ma2021pixel}, increasing accessibility for those with communicative impairments~\cite{johnson2018assessing,prnewswire2024deepbrain}, transforming education methods with interactive AI tutoring~\cite{bozkurt2023speculative,kessler2018technology}, and providing therapeutic support and social interaction in healthcare~\cite{rehm2016role,leff2014avatar}.

As one step towards achieving such capabilities, our work introduces \name, a new method that can produce audio-generated talking faces with a high level of realism and liveliness.
Given a static face image of an arbitrary individual, alongside a speech audio clip from any person, our approach is capable of generating a hyper-realistic talking face video efficiently. 
This video not only features lip movements that are meticulously synchronized with the audio input but also exhibits a wide range of natural, human-like facial dynamics and head movements.

Creating talking faces from audio has attracted significant attention in recent years with numerous approaches proposed \cite{zhou2020makelttalk,prajwal2020lip,zhang2021flow,sun2021speech2talking,guo2021ad,wang2021audio2head,wang2022one,wang2023lipformer,yu2023talking,zhang2023sadtalker,ma2023styletalk,he2024gaia}. 
However, existing techniques are still far from achieving the authenticity of natural talking faces. Current research has predominantly focused on the precision of lip synchronization with promising accuracy obtained \cite{prajwal2020lip,wang2023lipformer}. The creation of expressive facial dynamics and the subtle nuances of lifelike facial behavior remain largely neglected. 
This results in generated faces that seem rigid and unconvincing.
Additionally, natural head movements also play a vital role in enhancing the perception of realism.
Although recent studies have attempted to simulate realistic head motions~\cite{wang2021audio2head,yu2023talking,zhang2023sadtalker}, there remains a sizable gap between the generated animations and the genuine human movement patterns.

\begin{figure}[t!]
	\centering
	\includegraphics[width=.993\textwidth]{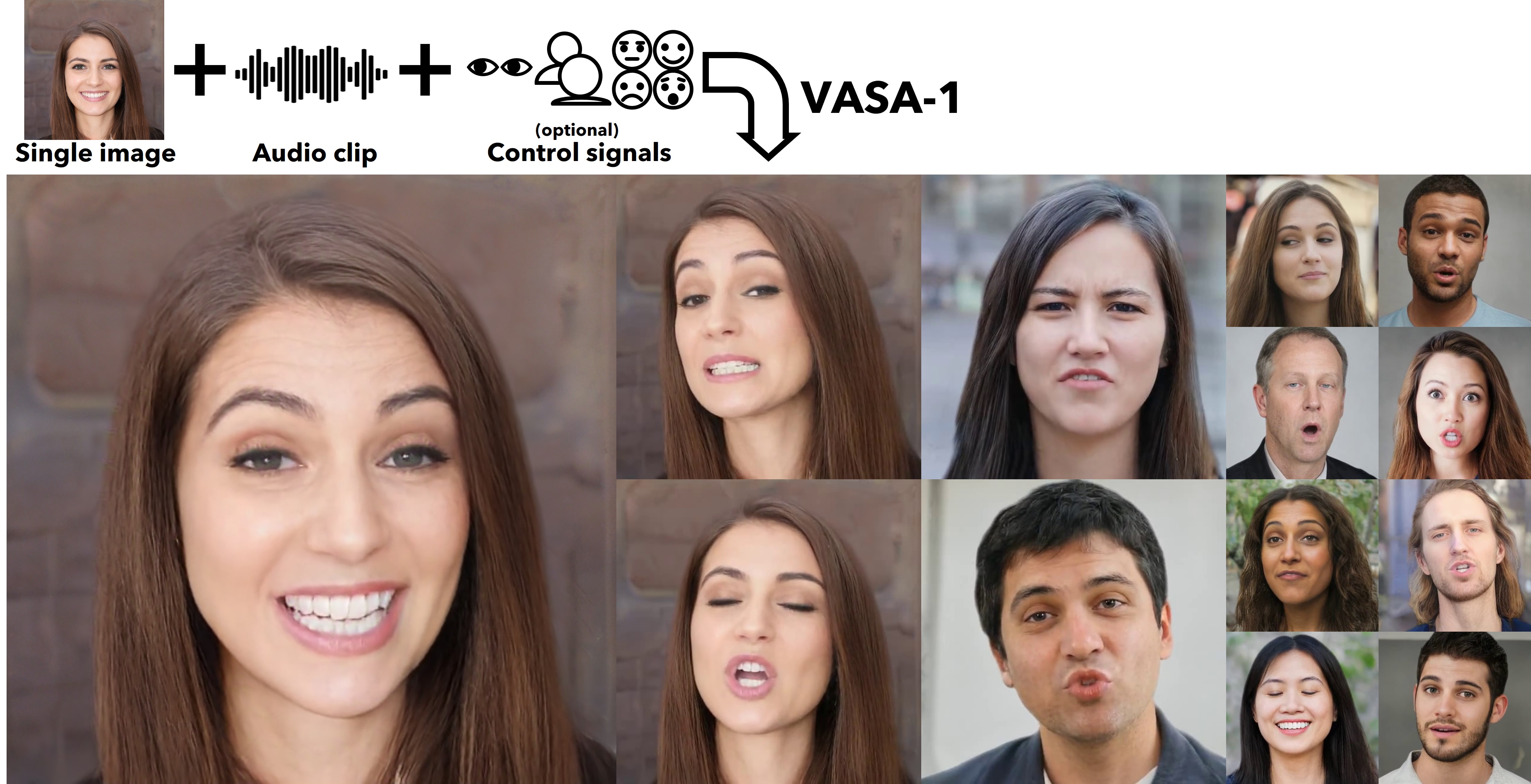}
	\vspace{-3pt}
	\caption{Given a single portrait image, a speech audio clip, and optionally a set of other control signals, our approach produces a high-quality lifelike talking face video of 512$\times$ 512 resolution at up to 40 FPS. The method is generic and robust, and the generated talking faces can faithfully mimic human facial expressions and head movements, reaching a high level of realism and liveliness. (\emph{All the photorealistic portrait images in this paper are virtual, non-existing identities generated by \cite{karras2020analyzing,betker2023improving}. See our project page for the generated video samples with audios.})}
	\vspace{-2pt}
	\label{fig:teaser}
\end{figure}

Another important factor is the efficiency of generation, which plays a pivotal role in real-time applications such as live communication. While image and video diffusion techniques have brought remarkable advancements in talking face generation~\cite{du2023dae,stypulkowski2024diffused,tian2024emo} as well as the broader video generation field~\cite{blattmann2023stable,videoworldsimulators2024}, the substantial computation demands have limited their practicality for interactive systems. A critical need exists for optimized algorithms that can bridge the gap between high-quality video synthesis and the low-latency requirements of real-time applications.

Given the limitations of existing methods, this work develops an efficient yet powerful audio-conditioned generative model that works in the \emph{latent space} of head and facial movements. Different from prior works, we train a Diffusion Transformer model on the latent space of \emph{holistic facial dynamics} as well as head movements. We consider all possible facial dynamics -- including lip motion, (non-lip) expression, eye gaze and blinking, among others -- as a single latent variable and model its probabilistic distribution in a unified manner. By contrast, existing methods often apply separate models for different factors, even with interleaved regressive and generative formulations for them~\cite{wang2021audio2head,zhou2021pose,yu2023talking,wang2023progressive,zhang2023sadtalker}. Our holistic facial dynamics modeling, together with the jointly learned head motion patterns, leads to the generation of a diverse array of lifelike and emotive talking behaviors.  
Furthermore, we incorporate a set of optional conditioning signals such as main gaze direction, head distance, and emotion offset into the learning process. This makes the generative modeling of complex distribution more tractable and increases the generation controllability. 

To achieve our goal, another challenge lies in constructing the latent space for the aforementioned holistic facial dynamics and gathering the data for the diffusion model training. Beyond facial and head movements, a human face image contains other factors such as identity and appearance. In this work, we seek to build a proper latent space for human face using a large volume of face videos. Our aim is for the face latent space to possess both a total state of \emph{disentanglement} between facial dynamics and other factors, as well as a high degree of \emph{expressiveness} to model rich facial appearance details and dynamic nuances.
We base our method on the 3D-aided representation~\cite{wang2021one,drobyshev2022megaportraits} which was proven to be expressive, and equip it with a collection of newly-designed loss functions critical to effective disentanglement. Without the new designs we can never reach a high quality of talking face generation, especially the liveliness with nuanced emotions. Trained on face videos in an self-supervised or weakly-supervised manner, our encoder can produce well-disentangled factors including 3D appearance, identity, head pose and holistic facial dynamics, and the decoder can generate high quality faces following the given latent codes.

\name has collectively advanced the realism of lip-audio synchronization, facial dynamics, and head movement to new heights. Coupled with high image generation quality and efficient running speed, we achieved real-time talking faces that are realistic and lifelike. 
Through detailed evaluations, we show that our method significantly outperforms existing methods on a set of  metrics, including a novel data-driven metric called Contrastive Audio and Pose Pretraining (CAPP) for measuring the audio-pose alignment and a pose variation intensity score that is related to the vividness of head motion. We believe \name brings us closer to a future where digital AI avatars can engage with us in ways that are as natural and intuitive as interactions with real humans, demonstrating appealing visual affective skills for more dynamic and empathetic information exchange.

\section{Related Work}

\paragraph{Disentangled face representation learning.} The representation of facial images through disentangled variables has been extensively studied by previous works. Some methods utilize sparse keypoints~\cite{siarohin2019first, zakharov2020fast} or 3D face models~\cite{ren2021pirenderer, gao2023high, zhang2023metaportrait} to explicitly characterize facial dynamics and other properties, but these can suffer from issues such as inaccurate reconstructions or limited expressive capabilities. There are also many works dedicated to learning disentangled representations within a latent space. A common approach involves separating faces into identity and non-identity components, then recombining them across different frames, either in a 2D~\cite{burkov2020neural,zhou2021pose, liang2022expressive,yin2022styleheat,pang2023dpe, wang2023progressive,tan2024edtalk} or 3D context~\cite{wang2021one, drobyshev2022megaportraits,drobyshev2024emoportraits}. The main challenge faced by these methods is the effective disentanglement of various factors while still achieving expressive representations of all static and dynamic facial attributes, which is addressed in this work.


\vspace{-5pt}
\paragraph{Audio-driven talking face generation.}
Talking face video generation from audio inputs has been a long-standing task in computer vision and graphics. 
Early works have focused on synthesizing only the lips, achieved by mapping audio signals directly to lip movements while leaving other facial attributes unchanged~\cite{suwajanakorn2017synthesizing, chen2018lip, prajwal2020lip, yin2022styleheat, cheng2022videoretalking}. More recent efforts have expanded the scope to include a broader array of facial expressions and head movements derived from audio inputs. For instance, the method of \cite{zhang2023sadtalker} separates the generation targets into different categories, including lip-only 3DMM coefficients, eye blinks, and head poses. \cite{yu2023talking} proposed to decompose lip and non-lip features on the top of the expression latent from \cite{zhou2021pose}. Both \cite{zhang2023sadtalker} and \cite{yu2023talking} regress lip-related representations directly from audio features and model other attributes in a probabilistic manner. In contrast to these approaches, our method generates comprehensive facial dynamics and head poses from audio along with other control signals. This approach differs from the trend of further disentanglement, seeking instead to create more holistic and integrated outputs.


\vspace{-5pt}
\paragraph{Video generation.} 
Recent advances in generative models~\cite{brown2020language, ho2020denoising, song2020score, song2020denoising} have led to significant progress in video generation. Earlier video generation approaches~\cite{vondrick2016generating, tulyakov2018mocogan, skorokhodov2022stylegan} employed the adversarial learning~\cite{goodfellow2014generative} framework, while more recent methods~\cite{yan2021videogpt, blattmann2023align, girdhar2023emu, kondratyuk2023videopoet, bar2024lumiere, videoworldsimulators2024} have leveraged diffusion or auto-regressive models to capture diverse video distributions.
Recently, several works concurrent to us \cite{tian2024emo, wei2024aniportrait} have adapted video diffusion techniques to audio-driven talking face generation, achieving promising results despite the slow training and inference speeds.
In contrast, our method is able to not only generating high-quality results but also achieve real-time efficiency -- a metric crucial to efficiency-demanding applications such as live communication.

\section{Method}

\paragraph{Overall framework.} As illustrated in Fig.~\ref{fig:teaser}, our method takes a single face image, optional control signals, and a speech audio clip to produce a realistic talking face video. Instead of generating video frames directly, we generate holistic facial dynamics and head motion in the latent space conditioned on audio and other signals. To achieve this, we start by constructing a face latent space and training the face encoder and decoder. An expressive and disentangled face latent learning framework is crafted and trained on real-life face videos. Then we train a simple yet powerful Diffusion Transformer to model the motion distribution and generate the motion latent codes in the test time given audio and other conditions.

\subsection{Expressive and Disentangled Face Latent Space Construction} \label{sec:latentspace}

Given a corpus of unlabeled talking face videos, we aim to build a latent space for human face with high degrees of \emph{disentanglement} and \emph{expressiveness}. 
The disentanglement enables effective generative modeling of the human head and holistic facial behaviors on massive videos, irrespective of the subject identities. It also enables disentangled factor control of the output which is desirable in many applications. Existing methods fall short of either   expressiveness~\cite{burkov2020neural,ren2021pirenderer,yu2023talking,wang2023progressive} or disentanglement~\cite{wang2021one,drobyshev2022megaportraits,zhang2023metaportrait} or both. The expressiveness of facial appearance and dynamic movements, on the other hand, ensures that the decoder can output high quality videos with rich facial details and the latent generator is able to capture nuanced facial dynamics.

To achieve this, we base our model on the 3D-aid face reenactment framework from \cite{wang2021one,drobyshev2022megaportraits}. The 3D appearance feature volume can better characterize the appearance details in 3D compared to 2D feature maps. The explicit 3D feature warping is also powerful in modeling 3D head and facial movements. 
Specifically, we decompose a facial image into a canonical 3D appearance volume $\mathbf{V}^{app}$, an identity code $\mathbf{z}^{id}$, a 3D head pose $\mathbf{z}^{pose}$, and a facial dynamics code $\mathbf{z}^{dyn}$. Each of them is extracted from a face image by an independent encoder, except that $\mathbf{V}^{app}$ is constructed by first extracting a posed 3D volume followed by rigid and non-rigid 3D warping to the canonical volume, as done in \cite{drobyshev2022megaportraits}. 
A single decoder $\mathcal{D}$ takes these latent variables as input and reconstructs the face image, where similar warping fields in the inverse direction are first applied to $\mathbf{V}^{app}$ to get the posed appearance volume. Readers are referred to \cite{drobyshev2022megaportraits} for more details of this architecture.

To learn the disentangled latent space, the core idea is to construct image reconstruction loss by swapping latent variables between different images in videos. Our basic loss functions are adapted from \cite{drobyshev2022megaportraits}. However, \emph{we identified the poor disentanglement between facial dynamics and head pose using the original losses. The disentanglement between identity and motions is also imperfect.} Therefore, we introduce several additional losses crucial to achieve our goal. Inspired by \cite{pang2023dpe}, we add a pairwise head pose and facial dynamics transfer loss to improve their disentanglement. Let $\mathbf{I}_i$ and $\mathbf{I}_j$ be two frames randomly sampled from the same video. We extract their latent variables using the encoders, and transfer $\mathbf{I}_{i}$'s head pose onto $\mathbf{I}_{j}$ as  $\hat{\mathbf{I}}_{j, \mathbf{z}^{pose}_{i}}=\mathcal{D}(\mathbf{V}^{app}_{j},\mathbf{z}^{id}_{j},\mathbf{z}^{pose}_{i},\mathbf{z}^{dyn}_{j})$ and $\mathbf{I}_{j}$'s facial motion onto $\mathbf{I}_{i}$ as $\hat{\mathbf{I}}_{i,\mathbf{z}^{dyn}_{j}}=\mathcal{D}(\mathbf{V}^{app}_{i},\mathbf{z}^{id}_{i},\mathbf{z}^{pose}_{i},\mathbf{z}^{dyn}_{j})$. The discrepancy loss $l_{consist}$ between $\hat{\mathbf{I}}_{j, \mathbf{z}^{pose}_{i}}$ and $\hat{\mathbf{I}}_{i,\mathbf{z}^{dyn}_{j}}$ is subsequently minimized. To reinforce the disentanglement between identity and motions, we add a face identity similarity loss $l_{cross\_id}$ for the cross-identity pose and facial motion transfer results. Let $\mathbf{I}_s$ and $\mathbf{I}_d$ be the video frames of two different subjects, we can transfer the motions of $\mathbf{I}_d$ onto $\mathbf{I}_s$ and obtain $\hat{\mathbf{I}}_{s,\mathbf{z}^{pose}_{d},\mathbf{z}^{dyn}_{d}}=\mathcal{D}(\mathbf{V}^{app}_{s},\mathbf{z}^{id}_{s},\mathbf{z}^{pose}_{d},\mathbf{z}^{dyn}_{d})$. Then, a cosine similarity loss between the deep face identity features~\cite{deng2019arcface} extracted from $\mathbf{I}_s$ and $\hat{\mathbf{I}}_{s,\mathbf{z}^{pose}_{d},\mathbf{z}^{dyn}_{d}}$ is applied. As we'll show in the experiments, our new loss function deigns are crucial to achieve an effective factor disentanglement and facilitate the high-quality, lifelike talking face generation.

\begin{figure}[t]
\vspace{-7pt}
	\centering
	\includegraphics[width=.99\textwidth]{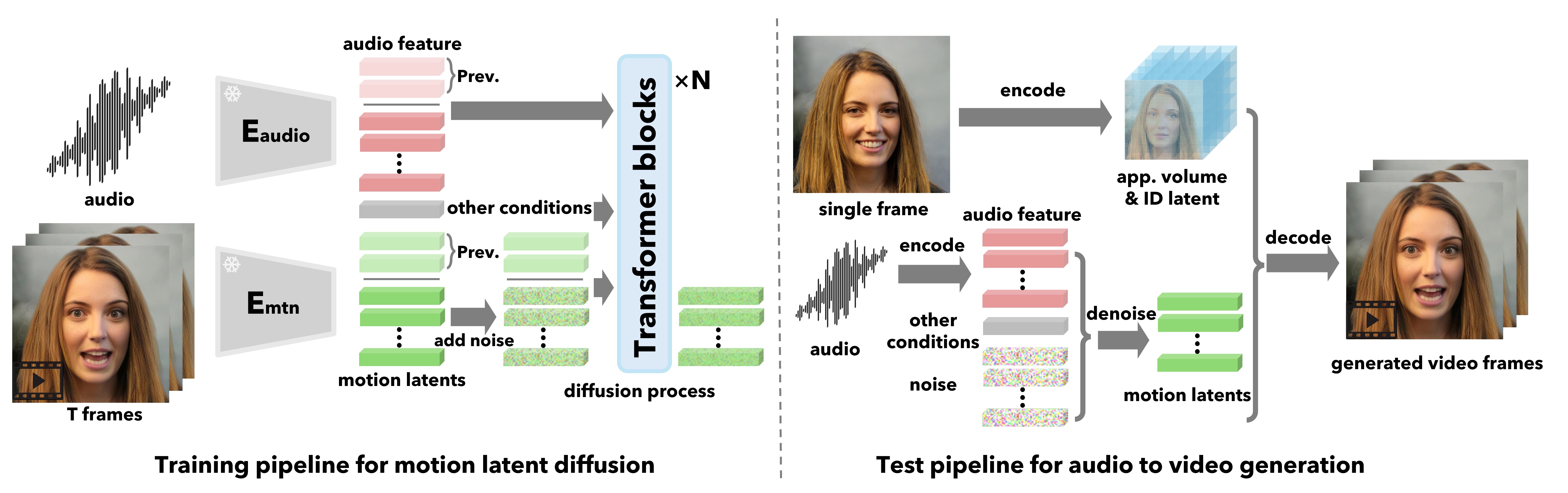}
	\vspace{-7pt}
	\caption{Our holistic facial dynamics and head pose generation framework with diffusion transformer.}
	\label{fig:pipeline}
 \vspace{-2pt}
\end{figure}

\subsection{Holistic Facial Dynamics Generation with Diffusion Transformer}
Given the constructed face latent space and trained encoders, we can extract the facial dynamics and head movements from real-life talking face videos and train a generative model. Crucially, we consider identity-agnostic holistic facial dynamics generation (HFDG), where our learned latent codes represent all facial movements such as lip motion, (non-lip) expression, and eye gaze and blinking. This is in contrast to existing methods that apply separate models for different factors with interleaved regression and generative formulations~\cite{wang2021audio2head,zhou2021pose,yu2023talking,wang2023progressive,zhang2023sadtalker}. Furthermore, previous methods often train on a limited number of identities~\cite{zhang2023sadtalker,xing2023codetalker,fan2022faceformer} and  cannot model the wide range of motion patterns of different humans, especially given an expressive motion latent space. 

In this work, we utilize diffusion models for audio-conditioned HFDG and train on massive talking face videos from a large number of identities. In particular, we apply a transformer architecture~\cite{vaswani2017attention,peebles2023scalable,sun2023diffposetalk} for our sequence generation task. Figure~\ref{fig:pipeline} shows an overview of our HFDG framework. 

Formally, a motion sequence extracted from a video clip is defined as $\mathbf{X} = \{[\mathbf{z}^{pose}_i,\mathbf{z}^{dyn}_i]\}, i=1,\ldots,W$. Given its accompanying  audio clip $\mathbf{a}$, we extract the synchronized audio features $\mathbf{A} = \{\mathbf{f}^{audio}_i\}$, for which we use a pretrained feature extractor Wav2Vec2~\cite{baevski2020wav2vec}.

\paragraph{Diffusion formulation.} Diffusion models define two Markov chains~\cite{ho2020denoising, song2020denoising, song2020score}, the forward chain progressively adds Gaussian noise to the target data, while the reverse chain iteratively restores the raw signal from noise. Following the denoising score matching objective~\cite{song2020score}, we define the simplified loss function as
\begin{equation}
\mathbb{E}_{t\sim \mathcal{U}[1, T],~\mathbf{X}^0,\mathbf{C}\sim q(\mathbf{X}^0, \mathcal{C})}(\|\mathbf{X}^0-\mathcal{H}(\mathbf{X}^t, t, \mathbf{C})\|^2),
\end{equation}
where $t$ denotes the time step, $\mathbf{X}^0 = \mathbf{X}$ is the raw motion latent sequence, and $\mathbf{X}^t$ is the noisy inputs generated by the diffusion forward process $q(\mathbf{X}^t|\mathbf{X}^{t-1}) = \mathcal{N}(\mathbf{X}^t;\sqrt{1 - \beta_t}\mathbf{X}^{t-1}, \beta_t\mathrm{I})$. $\mathcal{H}$ is our transformer network which predicts the raw signal itself instead of noise. $\mathbf{C}$ is the condition signal, to be described next.  

\paragraph{Conditioning signals.} The primary condition signal for our audio-driven motion generation task is the audio feature sequence $\mathbf{A}$. We also incorporate several additional signals, which not only make the generative modeling more tractable but also increase the generation controllability. 

Specifically, we consider the main eye gaze direction $\mathbf{g}$, head-to-camera distance $d$, and emotion offset $\mathbf{e}$. The main gaze direction, $\mathbf{g}=(\theta,\phi)$, is defined by a vector in spherical coordinates. It specifies the focused direction of the generated talking face.
We extract $\mathbf{g}$ for the training video clips using \cite{abdelrahman2023l2cs} on each frame followed by a simple histogram-based clustering algorithm. 
The head distance ${d}$ is a normalized scalar controlling the distance between the face and the virtual camera, which affects the face scale in the generated face video.
We obtain this scale label for the training videos using \cite{deng2019accurate}.
The emotion offset $\mathbf{e}$ modulates the depicted emotion on the talking face.
Note that emotion is often intrinsically linked to and can be largely inferred from  audio; hence, $\mathbf{e}$ serves only as a \emph{global offset} added to enhance or moderately alter the emotion when required. It is \emph{not} designed to achieve a total emotion shift during inference or produce emotions incongruent with the input audio. In practice, we use the averaged emotion coefficients extracted by \cite{savchenko2022hsemotion} as our emotion signal. 

In order to achieve a seamless transition between adjacent windows, we incorporate the last $K$ frames of the audio feature and generated motions from the previous window as the condition of the current one.  To summarize, our input condition can be denoted as $\mathbf{C}=[\mathbf{X}^{pre}, \mathbf{A}^{pre}; \mathbf{A}, \mathbf{g}, d, \mathbf{e}]$. All conditions are concatenated with noise along the temporal dimension as the input to the transformer. 

\paragraph{Classifier-free guidance (CFG)~\cite{ho2022classifier}.} In the training stage, we randomly drop each of the input conditions. During inference, we apply
\begin{eqnarray}
	\hat{\mathbf{X}}^0 = (1 + \sum_{\mathbf{c}\in\mathbf{C}}\lambda_\mathbf{c} )\cdot \mathcal{H}(\mathbf{X}^t, t, \mathbf{C}) 
	- \sum_{\mathbf{c}\in\mathbf{C}}\lambda_c  \cdot \mathcal{H}(\mathbf{X}^t, t, \mathbf{C}|_{\mathbf{c}=\emptyset})
	\label{eq:cfg}
\end{eqnarray}
where $\lambda_\mathbf{c}$ is the CFG scale for condition $\mathbf{c}$. $\mathbf{C}|_{\mathbf{c}=\emptyset}$ denotes that the condition $\mathbf{c}$ is replaced with $\emptyset$.

During training, we use a drop probability of $0.1$ for each condition except for $\mathbf{X}^{pre}$ and $\mathbf{A}^{pre}$ for which we use $0.5$. This is to ensure the model can well handle the first window with no preceding audio and motions (i.e., set to $\emptyset$). We also randomly drop the last few frames of $\mathbf{A}$ to ensure robust motion generation for audio sequences shorter than the window length.

\subsection{Talking Face Video Generation}

At inference time, given an arbitrary face image and an audio clip, we first extract the 3D appearance volume $\mathbf{V}^{app}$ and identity code $\mathbf{z}^{id}$ using our trained face encoders. Then, we extract the audio features, split them into segments of length $W$, and generate the head and facial motion sequences $\{\mathbf{X}=\{[\mathbf{z}^{pose}_i,\mathbf{z}^{dyn}_i]\}\}$ one by one in a sliding-window manner using our trained diffusion transformer $\mathcal{H}$. The final video can be generated subsequently using our trained decoder.

\vspace{-5pt}
\section{Experiments}\label{sec:experiments}
\paragraph{Implementation details.} 


For face latent space learning, we use the public VoxCeleb2 dataset from \cite{chung2018voxceleb2} which contains talking face videos from about 6K subjects. We reprocess the dataset and discard the clips with multiple individuals and those of low quality using the method of \cite{su2020blindly}. For motion latent generation, we use an 8-layer transformer encoder with an embedding dim $512$ and head number $8$ as our diffusion network. The model is trained on VoxCeleb2~\cite{chung2018voxceleb2} and another high-resolution talk video dataset collected by us, which contains about 3.5K subjects. In our default setup, the model uses a forward-facing main gaze condition, an average head distance of all training videos, and an empty emotion offset condition. The CFG parameters are set to $\lambda_\mathbf{A} = 0.5$ and $\lambda_\mathbf{g} = 1.0$, and $50$ sampling steps are used. Our face latent model takes around 7 days of training on a 4 NVIDIA RTX A6000 GPUs workstation, and the diffusion transformer takes around 3 days. The total data used for training comprises approximately 500K clips, each lasting between 2 to 10 seconds. The parameter counts of our 3D-aided face latent model and diffusion transformer model are about 200M and 29M respectively.

\vspace{-5pt}
\paragraph{Evaluation benchmarks.} We evaluate our method using two datasets. The first is a subset of  VoxCeleb2~\cite{chung2018voxceleb2}. We randomly selected 46 subjects from the test split of VoxCeleb2 and randomly sampled 10 video clips for each subject, resulting in a total of 460 clips. These video clips are about 5$\sim$15 seconds long (80\% are less than 10 seconds), with most of the content being interviews and news reports. To further evaluate our method under long speech generation with a wider range of vocal variations, we further collected 32 one-minute clips of 17 individuals. These videos are predominantly sourced from online coaching sessions and educational lectures and the talking styles are considerably more diverse than VoxCeleb2. We refer to this dataset as OneMin-32.
\vspace{-5pt}
\paragraph{Inference speed.} Our method generates video frames of 512$\times$512 size at 45fps in the offline batch processing mode, and can support up to 40fps in the online streaming mode with a preceding latency of only 170ms , evaluated on a desktop PC with a single NVIDIA RTX 4090 GPU.

\subsection{Quantitative Evaluation}

\paragraph{Evaluation metrics.} We use the following metrics for quantitative evaluation of our generated lip movement, head pose and overall video quality, including a new data-driven audio-pose synchronization metric trained in a way similar to CLIP~\cite{radford2021learning}:
\begin{itemize}[leftmargin=2em]
    \vspace{-2pt}
	\item \emph{Audio-lip synchronization.} We use a pretrained audio-lip synchronization network, i.e., SyncNet~\cite{chung2017out}, to assess the alignment of the input audio with the generated lip movements in videos. Specifically, we compute the confidence score and feature distance as $S_{C}$ and $S_{D}$ respectively. Higher $S_{C}$ and lower $S_{D}$  indicate better audio-lip synchronization quality in general.

	\item \emph{Audio-pose alignment.} Measuring the alignment between the generated head poses and input audio is not trivial and there are no well-established metrics. A few recent studies~\cite{zhang2023sadtalker,sun2023diffposetalk} employed the Beat Align Score~\cite{siyao2022bailando} to evaluate audio-pose alignment. However, this metric is not optimal because the concept of a ``beat'' in the context of natural speech and human head motion is ambiguous.
	In this work, we introduce a new data-driven metric called \emph{Contrastive Audio and Pose Pretraining (CAPP)} score. Inspired by CLIP~\cite{radford2021learning}, we jointly train a pose sequence encoder and an audio sequence encoder and predict whether the input pose sequence and audio are paired. The audio encoder is initialized from a pretrained Wav2Vec2 network~\cite{baevski2020wav2vec} and the pose encoder is a randomly initialized 6-layer transformer network. The input window size is 3 seconds. Our CAPP model is trained on 2K hours of real-life audio and pose sequences, and demonstrates a robust capability to assess the degree of synchronization between audio inputs and generate poses (see Sec.~\ref{sec:ablation}).

	\item \emph{Pose variation intensity.} We further define a pose variation intensity score $\Delta P$ which is the average of the pose angle differences between adjacent frames. 
    Averaged over all generated frames,
    $\Delta P$ provides an indication of the overall head motion intensity generated by a method.

	\item \emph{Video quality.} Following previous video generation works~\cite{yan2021videogpt, skorokhodov2022stylegan}, we use the Fréchet Video Distance (FVD)~\cite{unterthiner2019fvd} to evaluate the generated video quality. We compute the FVD metric using sequences of 25 consecutive frames, at resolution of 224$\times$224.
\end{itemize}

\paragraph{Compared methods.} 
We compare our method with there existing audio-driven talking face generation methods: MakeItTalk~\cite{zhou2020makelttalk},  Audio2Head~\cite{wang2021audio2head}, and SadTalker~\cite{zhang2023sadtalker}.

\begin{table}[t!]
	\centering
	\caption{Quantitative comparison with previous methods on two benchmarks. 
		}
    \vspace{-4pt}
	\resizebox{\textwidth}{!}{{\small 
		\begin{tabular}{c|cccc|ccccc}
			\hline
            & \multicolumn{4}{c|}{VoxCeleb2} & \multicolumn{5}{c}{OneMin-32} \\
            \hline
			&$S_C$ $\uparrow$ & $S_D$ $\downarrow$& \!CAPP$\uparrow$\! & $\Delta$P &$S_C$ $\uparrow$ & $S_D$ $\downarrow$& \!CAPP$\uparrow$\! & $\Delta$P& $\!\!\text{FVD}_{25}\downarrow$\!\!\\
			\hline
			MakeItTalk &4.176 &15.513 &-0.051 &0.210 & -0.123 &14.340 &0.002 &0.190 &304.83\\
   
			\!\!Audio2Head\!\! &6.172 &8.470 &0.246 &0.260 & 5.992 &8.211 &0.205 &0.239 &209.77\\
			SadTalker  &5.843 &8.813 &0.441 &0.275 & 5.501 &8.850 &0.383 &0.252 & 214.51 \\
			\emph{Ours} & $\bold{8.841}$ &$\bold{6.312}$ &$\bold{0.468}$ &$\bold{0.304}$ & $\bold{7.957}$ &$\bold{6.635}$ &$\bold{0.465}$ &$\bold{0.316}$ &\!\!$\bold{105.88}$\!\! \\
			\hline
            \textcolor{gray}{\emph{Ours~{\tiny (10\% data)}}} & \textcolor{gray}{8.818} &\textcolor{gray}{6.298} &\textcolor{gray}{0.457} &\textcolor{gray}{0.229} &\textcolor{gray}{7.990} &\textcolor{gray}{6.645} &\textcolor{gray}{0.441} &\textcolor{gray}{0.229} &\textcolor{gray}{\!\!{147.401}\!\!} \\
			\textcolor{gray}{Real video}  &\textcolor{gray}{7.640} &\textcolor{gray}{7.189} &\textcolor{gray}{0.588} &\textcolor{gray}{0.505} &\textcolor{gray}{7.192} &\textcolor{gray}{7.254} &\textcolor{gray}{0.559} &\textcolor{gray}{0.405} &\textcolor{gray}{29.25}\\
			\hline
		\end{tabular}
	}}
	\label{table:quantitative}
 \vspace{-10pt}
\end{table}

\paragraph{Main results.} For each audio input, we generate a single video for deterministic approaches, i.e., MakeItTalk and Audio2Head. For SadTalker and our method, we sample three videos for each audio and average the computed metrics.  
Since different pose representations are used by these methods, we re-extract the head poses from the generated frames to compute the pose-related metrics (i.e., CAPP and $\Delta P$).
For the FVD metric, we use 2K 25-frame video clips of both the real videos and generated ones. 
For reference purpose, we also report the evaluated metrics of real videos.

Table~\ref{table:quantitative} presents the results on the VoxCeleb2 and OneMin-32 benchmarks. Note that we did not evaluate the FVD on VoxCeleb2 as its video quality is varied and often low. On both benchmarks, our method achieves the best results among all methods on all evaluated metrics. In terms of audio-lip synchronization scores ($S_C$ and $S_D$), our method outperforms all others by a wide margin. Note that our method yields better scores than real videos, which is due to effect of the audio CFG (see Sec.~\ref{sec:ablation}). Our generated poses are better aligned with the audios especially on the OneMin-32 benchmark, as reflected by the CAPP scores. The head movements also exhibit the highest intensity according to $\Delta P$, although there's still a gap to the intensity of real videos. Our FVD score is significantly lower than others, demonstrating the much higher video quality and realism of our results.

\subsection{Qualitative Evaluation}

\begin{figure}[t!]
	\centering
	\includegraphics[width=.95\textwidth]{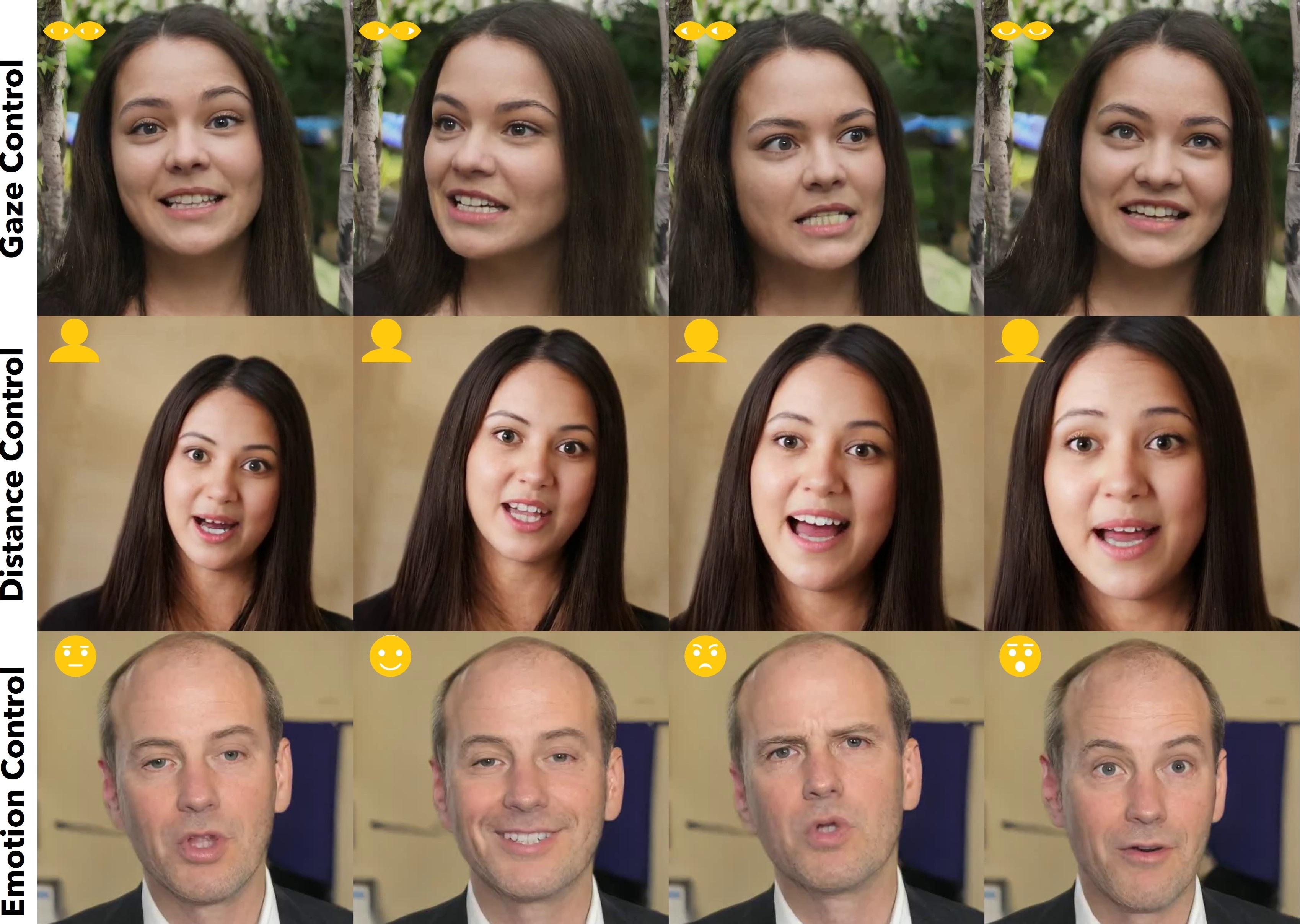}
	\vspace{-5pt}
	\caption{Generated talking faces under different control signals. \emph{Top row}: results under different main gaze direction condition (forward-facing, leftwards, rightwards, and upwards, respectively). \emph{Middle row}: results under different head distances (from far to near). \emph{Bottom row}: results under different emotion offset (neutral, happy, angry and surprised, respectively).
	}
\vspace{-10pt}
	\label{fig:controllability}
\end{figure}

\paragraph{Visual results.} Figure~\ref{fig:teaser} presents some representative audio-driven talking face generation results of our method. Visually inspected, our method can generate high-quality video frames with vivid facial emotions. Moreover, it can generate human-like conversational behaviors, including sporadic shifts in eye gaze during speech and contemplation, as well as the natural and variable rhythm of eye blinking, among other nuances. \emph{We highly recommend that readers view our video results in the supplementary material to fully perceive the capabilities and output quality of our method.}

\vspace{-5pt}
\paragraph{Generation controllability.} Figure~\ref{fig:controllability} shows our generated results under different control signals including main eye gaze, head distance, and emotion offset. Our  model can well interpret these signals and produce talking face results that closely adhere to these specified parameters.

\vspace{-5pt}
\paragraph{Disentanglement of face latents.} 
Figure~\ref{fig:id_exp} shows 
that when applying the same motion latent sequences onto different subjects, our method effectively maintains both the distinct facial movements and the unique facial identities. This indicates the efficacy of our method in disentangling identity and motion. 
Figure~\ref{fig:pose_exp} further illustrates the effective disentanglement between head pose and facial dynamics. By holding one aspect constant and changing the other, the resulting images faithfully reflect the intended head and facial motions without interference.

\vspace{-5pt}
\paragraph{Out-of-distribution generation.}  Our method exhibits the capability to handle photo and audio inputs that fall outside the training distribution, such as artistic photos, singing audio clips, and non-English speech, as illustrated in Figure~\ref{fig:ood}.

\paragraph{Comparison with other methods.} 
Some visual examples from different methods are presented in Figure~\ref{fig:push_ups}~\ref{fig:excruciating}~\ref{fig:what}~\ref{fig:lotsofquestions}. Our method outperforms the others in terms of the precise audio-lip synchronization and delivers  much more vivid and natural facial dynamics and head movements.

\subsection{Analysis and Ablation Study}\label{sec:ablation}

\paragraph{CAPP metric.} We analyze the effectiveness of our proposed CAPP metric in measuring the alignment between audio and head pose.

\setlength{\intextsep}{-6pt}
\begin{wraptable}[4]{r}{0.44\textwidth}
	\centering
	\caption{CAPP under frame shifting}\label{tab:capp_shift}
 {\small
	\begin{tabular}{ccccc}
		\hline
		0  & $\pm$1 &$\pm$2 & $\pm$3 &$\pm$4 \\
		\hline
		\!\textbf{0.608}\! &\!0.462\! &\!0.206\!  &\!0.069\! & \!0.082\!\\
		\hline
	\end{tabular}
 }
\end{wraptable}  First, we study its sensitivity to temporal shifting by manually introducing frame offsets to ground-truth audio-pose pairs. We extract 3-second clip segments from the VoxCeleb2 test split, yielding approximately 2.1K audio-pose pairs. The average CAPP score for these pairs is $0.608$, as shown in Table~\ref{tab:capp_shift}.
Manual frame shifts lead to a rapid decline in CAPP scores, approaching zero for shifts larger than two frames. This indicates a robust correlation between CAPP scores and audio-head pose alignment.

\begin{wraptable}[4]{r}{0.44\textwidth}
	\centering
	\caption{CAPP under pose variation scaling}\label{tab:capp_scale}
 {\small
	\begin{tabular}{cccccc}
		\hline
		$\times$0.2 &$\times$0.5  &$\times$1.0 & $\times$1.5 & $\times$3.0 \\
		\hline
		\!0.368\! &\!0.584\! & \!\textbf{0.608}\! &\!0.587\! &\!0.505\! \\
		\hline
	\end{tabular}
 }
\end{wraptable}
We further investigate the effect of head movement intensity on CAPP by manually scaling the pose differences between consecutive frames using various factors. Table~\ref{tab:capp_scale} shows that altering movement intensity negatively impacts the CAPP scores, demonstrating CAPP can assess the alignment of audio and pose in terms of their intensity. However, this sensitivity to intensity appears less pronounced than that to temporal misalignment.

\vspace{-6pt}
\paragraph{CFG scales for diffusion model.} 
The CFG strategy~\cite{ho2022classifier} for diffusion models can attain a trade-off between sample quality and diversity. Here we evaluate the choice of the CFG scales for the audio and main gaze conditions (i.e., $\lambda_\mathbf{A}$ and $\lambda_\mathbf{g}$ in Eq.~\ref{eq:cfg}) in our model.

As shown in Table~\ref{table:cfg}, as we increase the value of $\lambda_\mathbf{g}$, the accuracy of gaze control improves. Increasing the audio CFG scale to $\lambda_\mathbf{A}=0.5$ significantly enhances the performance of lip-audio alignment ($S_C$ and $S_D$), pose-audio alignment (CAPP), and pose variation intensity ($\Delta P$). With positive audio CFG, the lip-audio alignment scores even surpass those evaluated on real videos (the results without audio CFG, i.e., $\lambda_\mathbf{A}=0$, were slightly worse than or comparable to them). Moreover, the FVD score shows a slight drop which indicates slightly better video quality.

\begin{table}[t!]
	\small{
		\centering
		\caption{Ablation study of the audio and main gaze CFG scales as well as the sampling steps. $\mathcal{E}_g$ denotes the average angular error of main gaze directions and $\mathcal{E}_s$  is the average head distance error.}
		\begin{tabular}{lccccccc}
			\hline
			&$S_C$ $\uparrow$ & $S_D$ $\downarrow$ & CAPP$\uparrow$ & $\Delta$P& $\text{FVD}_{25}\downarrow$ &$\mathcal{E}_{\theta_g}\downarrow$&$\mathcal{E}_{s}\downarrow$\\
			\hline
			$\lambda_\mathbf{A}=0.0,  \ \!\lambda_\mathbf{g}=0.0$ &7.087 &7.391 &0.414 &0.291 &117.425 & 5.730 &0.004
			\\
			$\lambda_\mathbf{A}=0.0,  \ \!\lambda_\mathbf{g}=1.0$ &7.134 &7.345 &0.421 &0.290 &116.547 & 5.329 &0.004
			\\
			$\lambda_\mathbf{A}=0.0,  \ \!\lambda_\mathbf{g}=2.0$ &7.108 &7.386 &0.414 &0.298 &117.784 & 5.064 &0.005
			\\
			$\bold{\lambda_{A}=0.5, \lambda_\mathbf{g}=1.0}$ &7.957 &6.635 &0.465 &0.316 &105.884 & 5.253 &0.005
			\\
			$\lambda_\mathbf{A}=1.0,  \ \!\lambda_\mathbf{g}=1.0$ &8.218 &6.437 &0.474 &0.342 &104.886 &5.333 &0.005
			\\
			$\lambda_\mathbf{A}=2.0, \ \!\lambda_\mathbf{g}=1.0$ &8.295 &6.397 &0.455 & 0.395 &104.293 &5.531 &0.005
			\\
			\hline
			$\bold{\lambda_\mathbf{A}=0.5, \lambda_\mathbf{g}=1.0}$ {\footnotesize{(steps$=10$)}} &8.293 &6.363 &0.523 &0.243 &117.060 & 5.469 &0.006
			\\
			\hline
			\textcolor{gray}{Real video}  &\textcolor{gray}{7.192} &\textcolor{gray}{7.254} &\textcolor{gray}{0.559} &\textcolor{gray}{0.405} &\textcolor{gray}{29.244}&--&--\\
			\hline
		\end{tabular}
        
	\label{table:cfg}
 \vspace{-5pt}
	}
\end{table}

Further increasing $\lambda_\mathbf{A}$ marginally improves lip-audio synchronization and reduces $\text{FVD}_{25}$, but at the cost of slightly degrading audio-pose synchronization and gaze controllability. 
In addition, observations from the generated videos indicate that a higher $\lambda_\mathbf{A}$ significantly amplifies mouth movements for strong vocals and causes head pose jitter during rapid speech. For balanced performance and overall generation quality, we set  $\lambda_\mathbf{A}=0.5$ and $\lambda_\mathbf{g}=1.0$ as our standard configuration.


We also evaluated the influence of sampling steps on performance. Table~\ref{table:cfg} illustrates that decreasing the steps from $50$ to $10$ improves audio-lip and audio-pose alignment while compromising pose variation intensity and overall video quality. This step reduction could accelerate the inference process by a factor of 5 for this latent motion generation module.

\begin{figure}[t!]
	\centering
	\includegraphics[width=.99\textwidth]{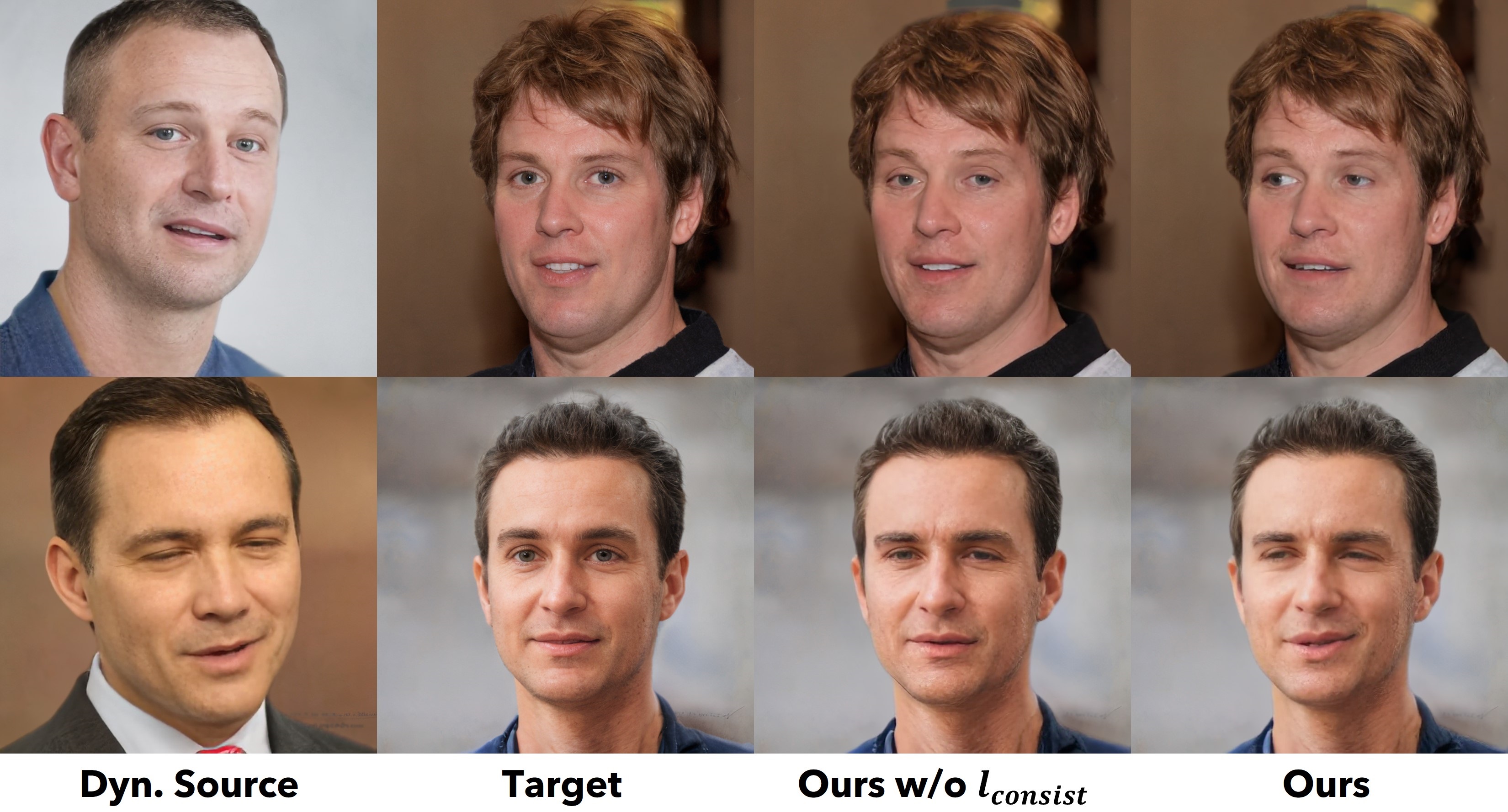}
	\vspace{-5pt}
	\caption{Ablation study on loss function $l_{consist}$ for disentangled latent space learning. We generate the results by only transferring the facial dynamics from source to target with head pose unchanged. $l_{consist}$ is crucial for decoupling  subtle yet important facial dynamics from head pose.
	}
	\label{fig:exp_consist}
    \vspace{-10pt}
\end{figure}

\paragraph{Training data scale.} To validate the data scale influence and compare our model with previous methods at similar scales, we trained a diffusion model using only 10\% of the data (i.e., 50K clips). As shown in Table~\ref{table:quantitative}, the model trained on this reduced dataset demonstrates comparable audio-lip and audio-pose synchronization to the full-dataset model, although the FVD and $\Delta$p metrics are not as good. Nonetheless, it still significantly outperforms previous methods across all metrics assessing synchronization, motion intensity, and video quality.
This indicates that our approach remains highly effective even with much less data, and that increasing the dataset size enhances motion diversity.


\vspace{-5pt}
\paragraph{Losses for latent space learning.} As described in Sec.~\ref{sec:latentspace}, we introduce new losses $l_{consist}$ and $l_{cross\_id}$ to improve the disentanglement of facial dynamics, head pose, and face identity. To validate the effectiveness of $l_{consist}$, we transfer only facial dynamics from the source image to the target while keeping the target's pose unchanged.
Figure~\ref{fig:exp_consist} shows that without $l_{consist}$, the latent model may struggle to replicate subtle facial dynamics such as side glances and lip asymmetries which are oftentimes coupled with head poses (e.g., a skewed mouth may coincide with a tilted head, and the gaze direction usually aligns with the head's pose). Decoupling these subtle yet important dynamics are challenging without explicit constraints from $l_{consist}$.

We also evaluate the face identity loss $l_{cross\_id}$ for cross identity driving during training. We use all 108 subjects from the VoxCeleb2 test set for evaluation. For each subject, we chose the image that is closest to a frontal view from the first frames of all its clips to serve as the target image. Then we randomly selected 50 clips of other subjects as source videos, which leads to a total of 5,400 cross-reenactment clips. 
We calculate the facial identity preservation score by averaging the facial identity feature cosine similarity over all generated frames of all subjects. With the introduced face identity loss $l_{cross\_id}$, this identity preservation score of our results improved from $0.72$ to $0.80$.

\vspace{-5pt}
\section{Conclusion}\label{sec:conclusion}
\vspace{-5pt}
In summary, our work presents an audio-driven talking face generation model that stands out for its efficient generation of realistic lip synchronization, vivid facial expressions, and naturalistic head movements from a single image and audio input. It significantly outperforms existing methods in delivering video quality and performance efficiency, demonstrating promising visual affective skills in the generated face videos.  The technical cornerstone is an innovative holistic facial dynamics and head movement generation model that works in an expressive and disentangled face latent space.

The advancements made by \name have the potential to reshape human-human and human-AI interactions across various domains, including communication, education, and healthcare. The integration of controllable conditioning signals further enhances the model's adaptability for personalized user experiences.

There are still several limitations with our method. Currently, it processes human regions only up to the torso. Extending to the full upper body could offer additional capabilities. While utilizing 3D latent representations, the absence of a more explicit 3D face model such as \cite{wu2022anifacegan,wu2023aniportraitgan} may result in artifacts like texture sticking due to the neural rendering. Additionally, our approach does not account for non-rigid elements like hair and clothing, which could be addressed with a stronger video prior. In the future, we also plan to incorporate more diverse talking styles and emotions to improve expressiveness and control.

\section*{Contribution statement}
Sicheng Xu, Guojun Chen, Yu-Xiao Guo were the core contributors to the implementation, training, and experimentation of various algorithm modules, as well as the data processing and management. Jiaolong Yang initiated the project idea, led the project, designed the overall framework, and provided detailed technical advice to each component. Chong Li, Zhengyu Zang and Yizhong Zhang contributed to enhancing the system quality, conducting evaluations, and demonstrating results. Xin Tong provided technical advice throughout the project and helped with project coordination. Baining Guo offered strategic research direction guidance, scientific advising, and other project supports. Paper written by Jiaolong Yang and Sicheng Xu.

\section*{Acknowledgments}
We would like to thank our colleagues Zheng Zhang, Zhirong Wu, Shujie Liu, Dong Chen, Xu Tan, Yu Deng, Lidong Zhou, and others for the valuable discussions and insightful suggestions for our project.

{
	\small
	\bibliographystyle{ieee_fullname}
	\bibliography{bib}
}

\newpage
\appendix
\setcounter{figure}{0}    
\renewcommand{\thefigure}{A.\arabic{figure}}


\vspace{-5pt}
\section{Societal Impacts and Responsible AI Considerations }\label{sec:rai}
\vspace{-5pt}
Our research focuses on generating audio-driven visual affective skills for virtual AI avatars, aiming for positive applications. 
It is not intended to create content that is used to mislead or deceive. However, like other related content generation techniques, it could still potentially be misused for impersonating humans. We are opposed to any behavior that creates misleading or harmful contents of real persons. Currently, the videos generated by this method still contain identifiable artifacts, and the numerical study in Section~\ref{sec:experiments} shows that there's still a gap to achieve the authenticity of real videos. Furthermore, we have trained a neural network based detector to distinguish real videos and those generated by our \name, and the detector shows a $97.8\%$ accuracy for this task.

While acknowledging the possibility of misuse, it's imperative to recognize the substantial positive potential of our technique. The benefits -- ranging from enhancing educational equity, improving accessibility for individuals with communication challenges, and offering companionship or therapeutic support to those in need -- underscore the importance of our research and other related explorations. We are dedicated to developing AI responsibly, with the goal of advancing human well-being.


To combat potential misuse of our technique and other related ones and provide necessary safeguards, we are also working on applying our method for advancing face media forgery detection. Specifically, we are training generic face forgery detection models that incorporate our generated talking face videos as part of the training data. Our preliminary exploration shows that using our method to generate training data can lead to an obvious improvement of generality for the forgery detection models, and we’ll keep the community updated on new progresses.

\section{More Qualitative Evaluation, Comparison and Ablation Study}
\begin{figure}[t!]
	\centering
	\includegraphics[width=.99\textwidth]{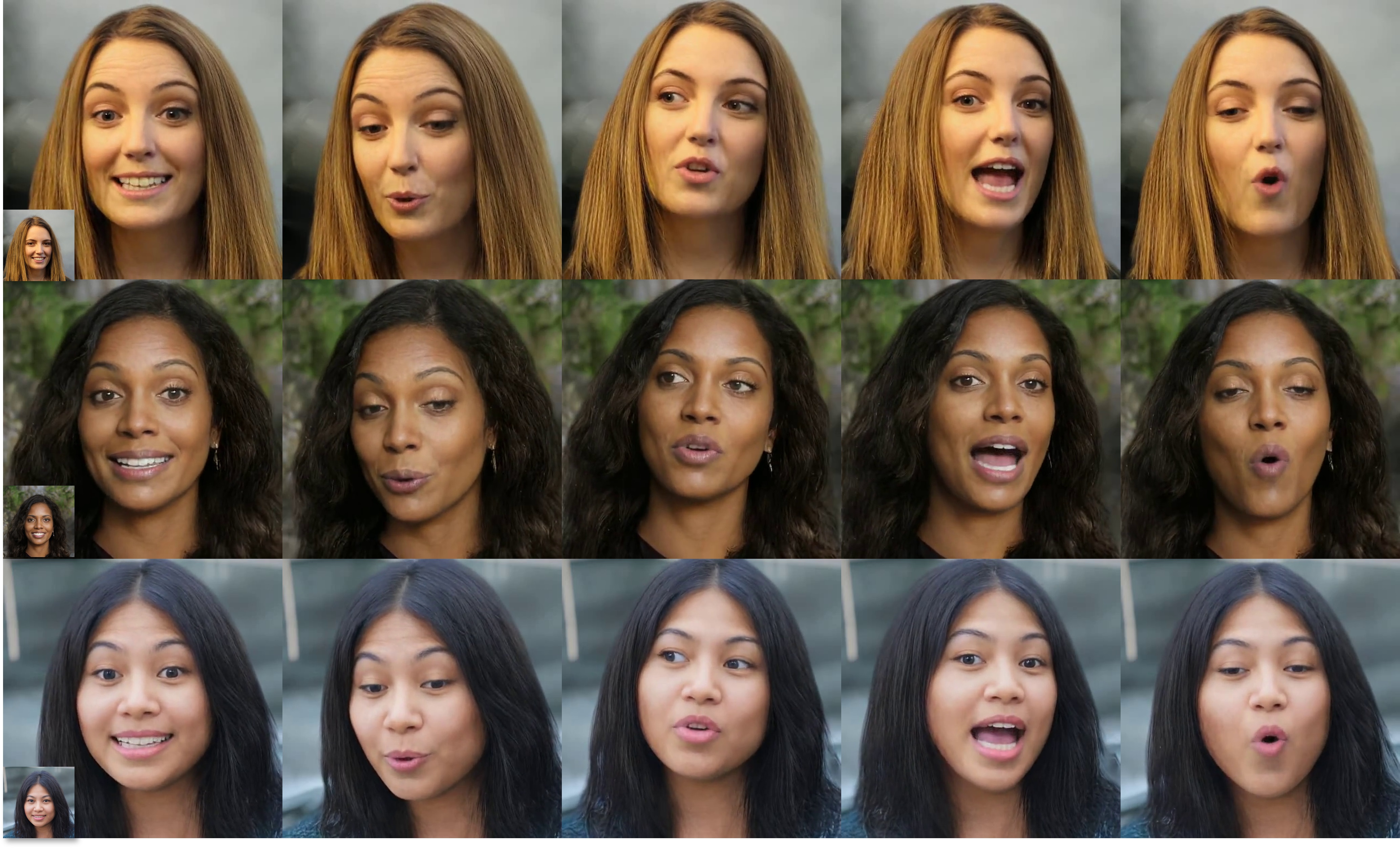}
	\vspace{-8pt}
	\caption{Disentanglement between identity and motion. In these examples, the same generated  head and facial motion sequences are applied onto three different face images.}
	\label{fig:id_exp}
\end{figure}

\begin{figure}[t!]
	\centering
	\includegraphics[width=.99\textwidth]{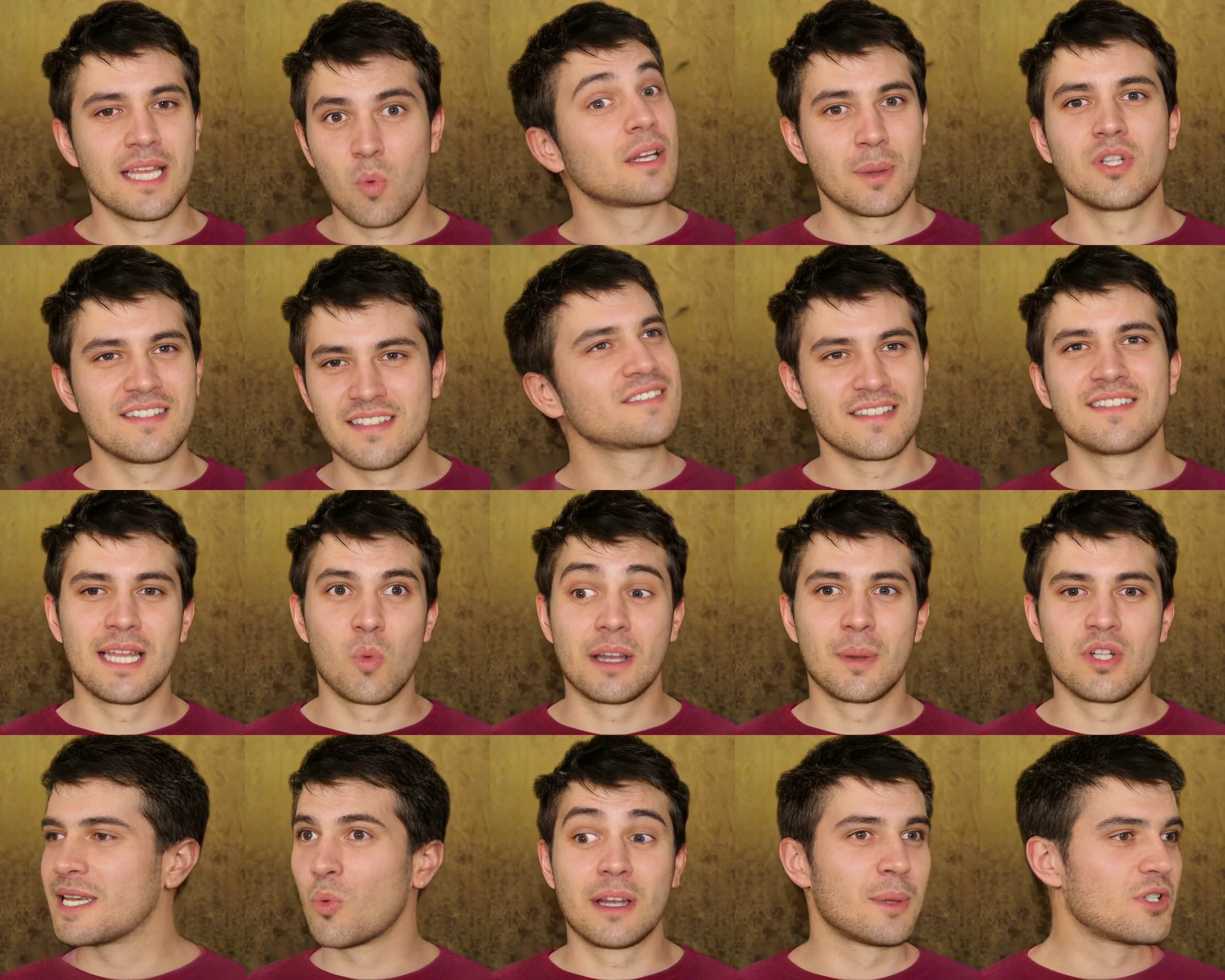}
	\vspace{-6pt}
	\caption{Disentanglement between head pose and facial dynamics. \emph{From top to bottom}: the raw generated sequence, applying generated poses with fixed initial facial dynamics, and applying generated facial dynamics with fixed initial head pose and pre-defined spinning poses, respectively.
	}
	\label{fig:pose_exp}
\end{figure}

\begin{figure}[t!]
	\centering
	\includegraphics[width=.99\textwidth]{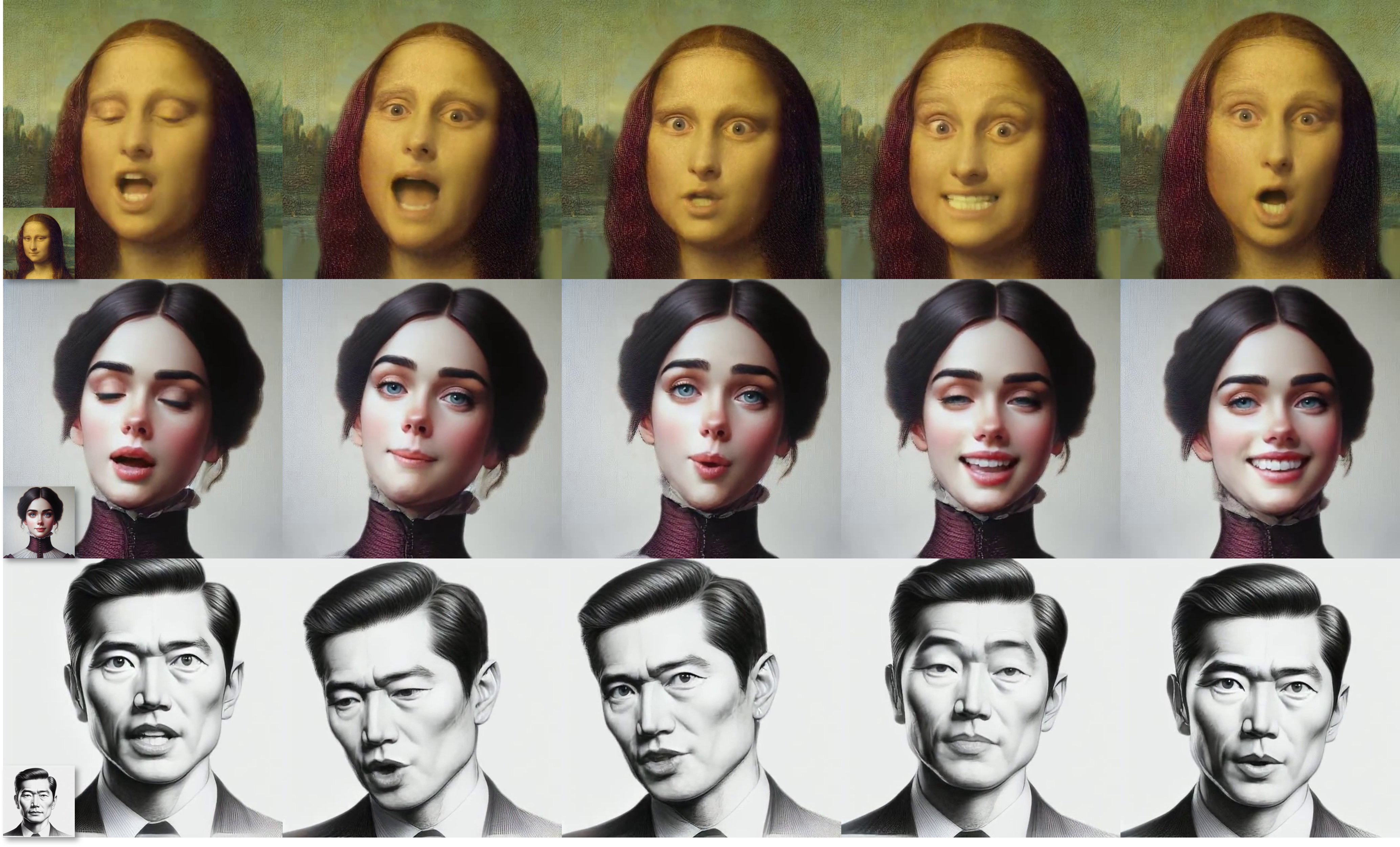}
	\vspace{-8pt}
	\caption{Generation results with out-of-distribution images (non-photorealistic) and audios (singing audios for the first two rows and non-English speech for the last row). Our method can still generate high quality videos well-aligned with the audios, although it was not trained on such data variations. 
    See the supplementary video with audio for a better illustration of these results.
	}
	\label{fig:ood}
\end{figure}

\begin{figure}[t!]
	\centering
	\includegraphics[width=1\textwidth]{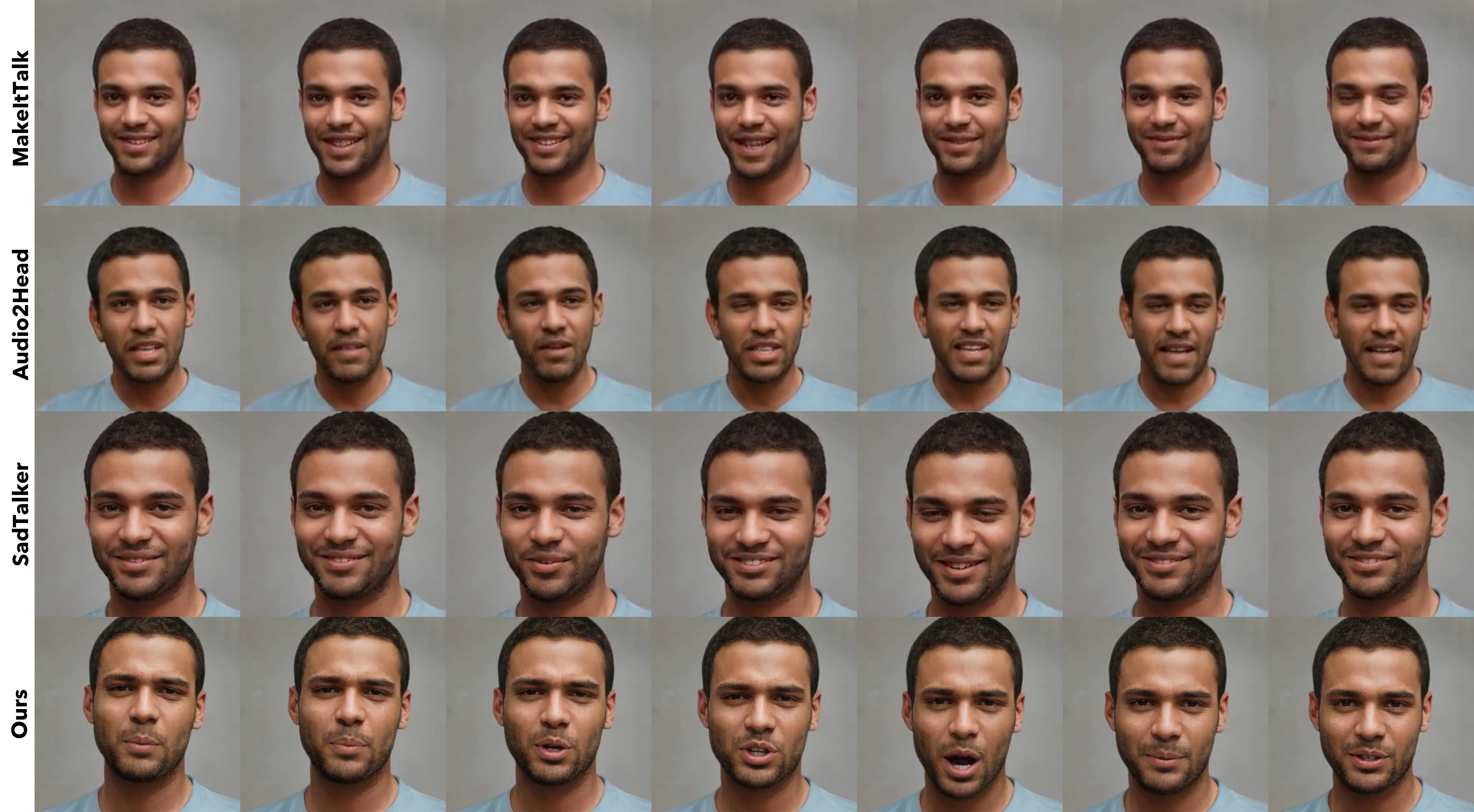}
	\vspace{-16pt}
	\caption{Generation results from different methods with the input audio segment uttering ``\texttt{push ups}''. See our supplementary video for a better illustration and comparison.
	}
	\label{fig:push_ups}
\end{figure}

\begin{figure}[t!]
	\centering
	\includegraphics[width=1\textwidth]{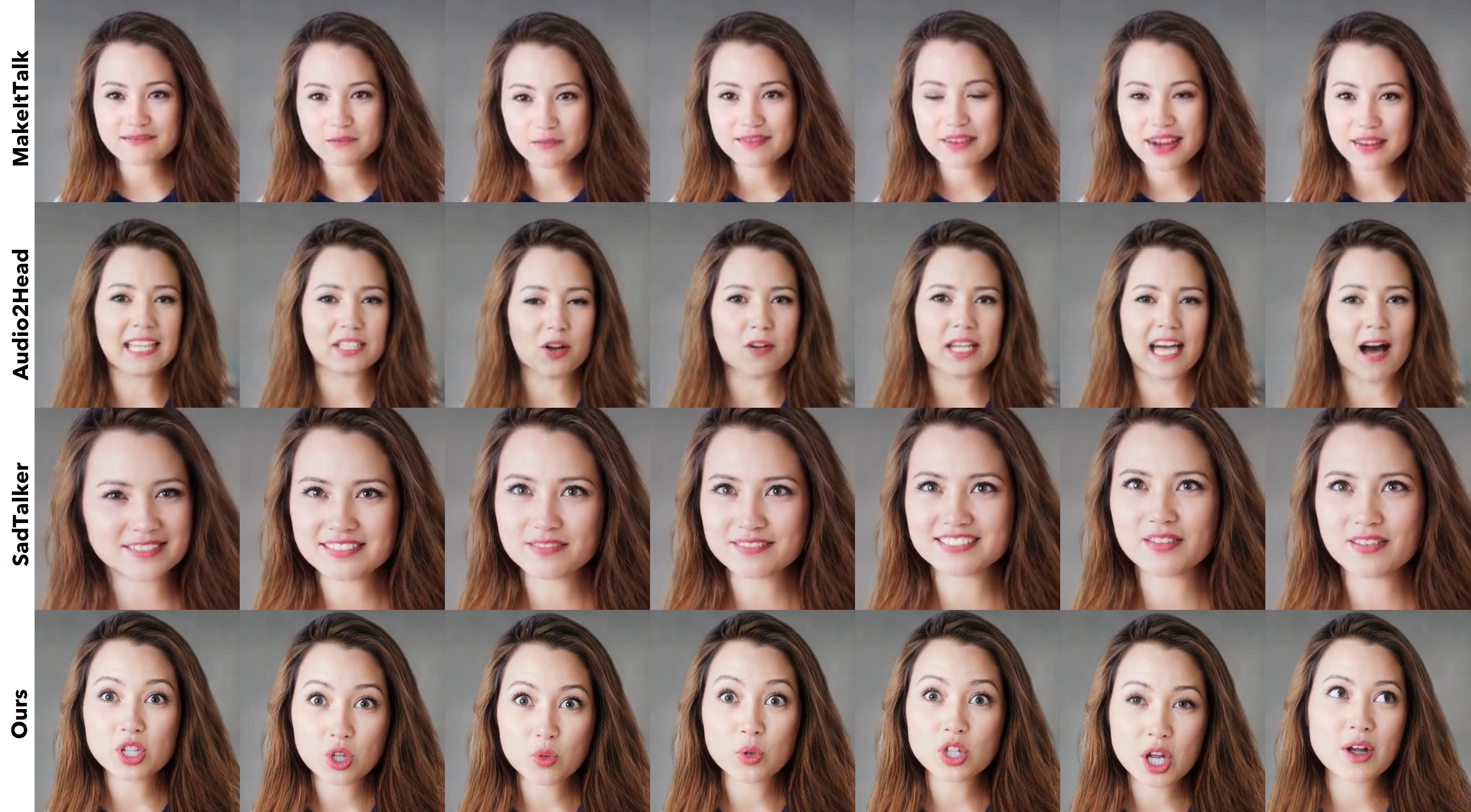}
	\vspace{-16pt}
	\caption{Generation results from different methods with the input audio segment uttering ``\texttt{excruciating}''. See our supplementary video for a better illustration and comparison.
	}
	\label{fig:excruciating}
\end{figure}

\begin{figure}[t!]
	\centering
	\includegraphics[width=1\textwidth]{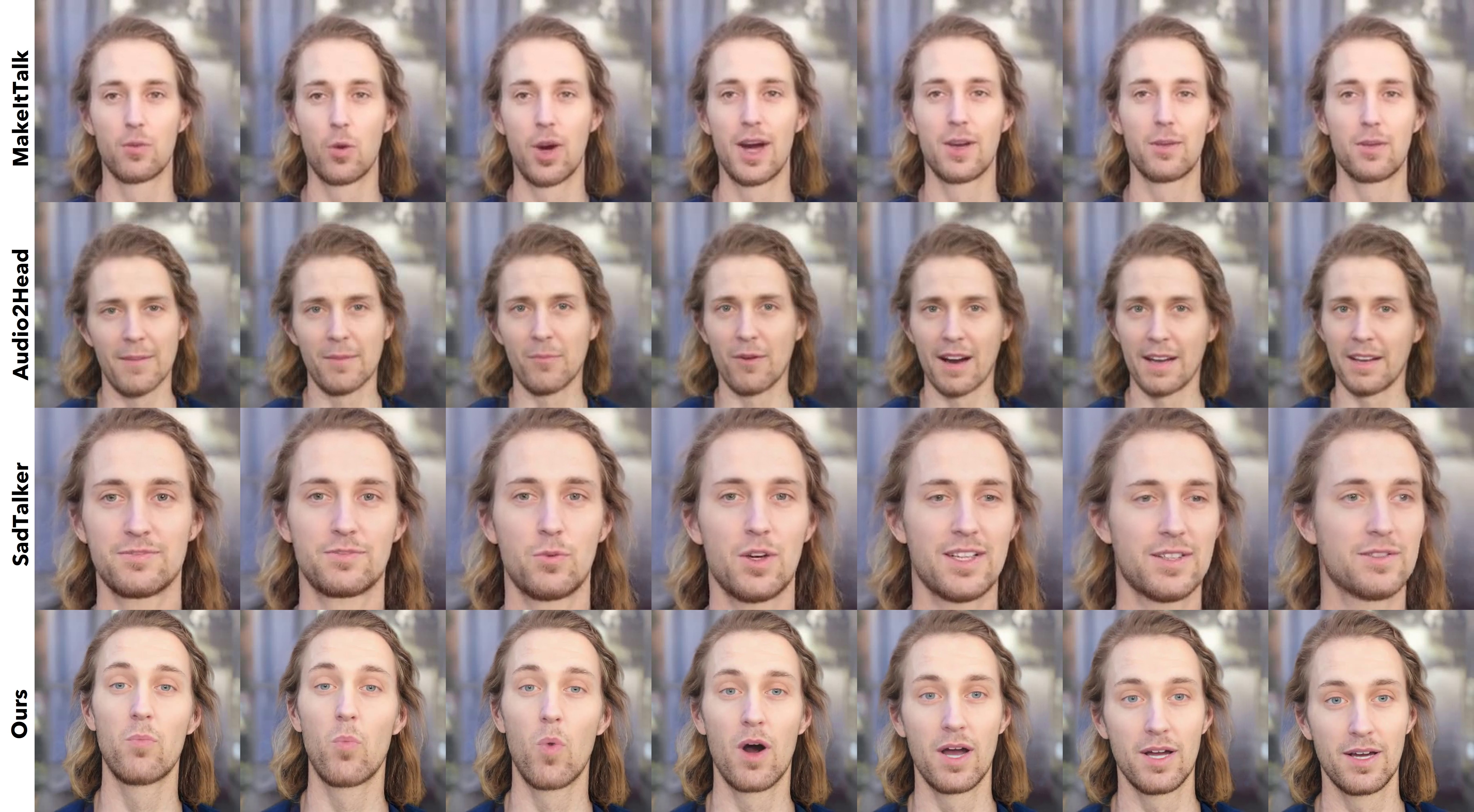}
	\vspace{-16pt}
	\caption{Generation results from different methods with the input audio segment uttering ``\texttt{what?}''. See our supplementary video for a better illustration and comparison.
	}
	\label{fig:what}
\end{figure}

\begin{figure}[t!]
	\centering
	\includegraphics[width=1\textwidth]{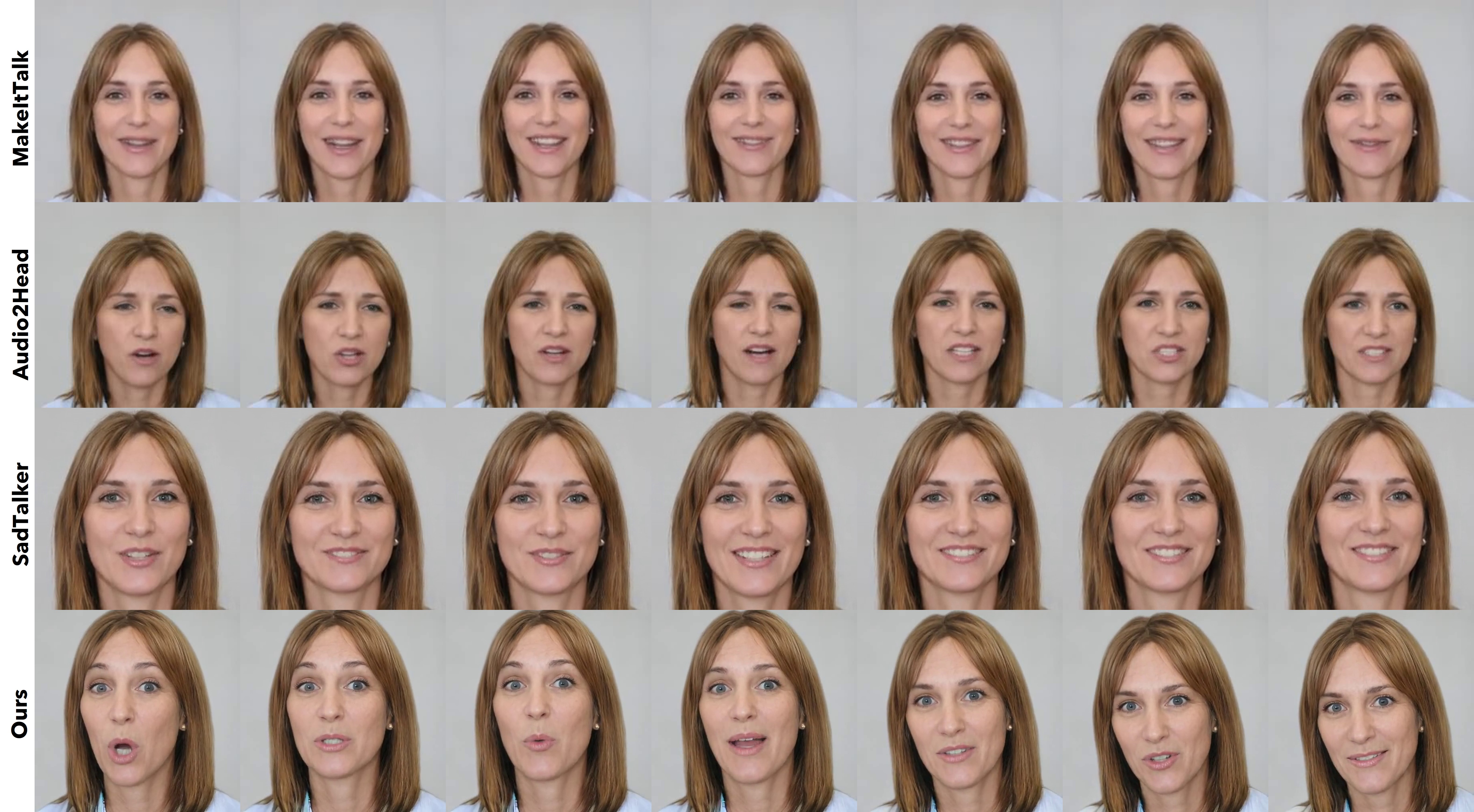}
	\vspace{-15pt}
	\caption{Generation results from different methods with the input audio segment uttering ``\texttt{lots of questions}''. See our supplementary video for a better illustration and comparison.
	}
\label{fig:lotsofquestions}
\end{figure}

See Figure~\ref{fig:id_exp}~\ref{fig:pose_exp}~\ref{fig:ood}~\ref{fig:push_ups}~\ref{fig:excruciating}~\ref{fig:what}~\ref{fig:lotsofquestions}.

\newpage

\end{document}